\documentclass{ngsm2024} 

\title[Activation Bottleneck]{Activation Bottleneck: Sigmoidal Neural Networks Cannot Forecast a Straight Line}

\optauthor{%
\Name{Maximilian Toller} \Email{mtoller@know-center.at}\\
\Name{Bernhard C Geiger} \Email{bgeiger@know-center.at}\\
\addr{Know-Center GmbH, Austria, EU}
\AND
\Name{Hussain Hussain} \Email{hhussain@tugraz.at}\\
\addr Graz University of Technology, Austria, EU
}

\newcommand{\R}{\mathbb{R}}
\newcommand{\D}{\mathbb{D}}
\usepackage{bm}
\begin{document}

\maketitle

\begin{abstract}%
A neural network has an activation bottleneck if one of its hidden layers has a bounded image.
We show that networks with an activation bottleneck cannot forecast unbounded sequences such as straight lines, random walks, or any sequence with a trend: The difference between prediction and ground truth becomes arbitrary large, regardless of the training procedure.
Widely-used neural network  architectures such as LSTM and GRU suffer from this limitation.
In our analysis, we characterize activation bottlenecks and explain why they prevent sigmoidal networks from learning unbounded sequences.
We experimentally validate our findings and discuss modifications to network architectures which mitigate the effects of activation bottlenecks.

\end{abstract}


\section{Introduction}
Since the early 2000s, it has been known that multi-layer perceptrons (MLPs) with sigmoidal activation functions are ``doomed to fail''~\citep{zhang2005neural} at forecasting sequences with trend. 
This insight can be explained on an intuitive level: The output of a sigmoidal function is always between 0 and 1, so the output of the final (linear) layer cannot be larger than the sum of its weights and bias.
However, a sequence with trend can become arbitrarily large.
Due to this shortcoming, MLPs have lost popularity for modeling sequences, and recurrent architectures such as Long Short-Term Memory cells (LSTMs)~\citep{hochreiter1997long} and Gated-Recurrent Units (GRUs)~\citep{cho2014learning} have become more popular.
The key idea of recurrent architectures is that they have a ``memory'' unit which keeps track of past states/values, which presumably allows them to model time-dependent sequences while simultaneously avoiding the \textit{vanishing gradient problem}~\citep{hochreiter1991untersuchungen}.
However, despite many successes, it remains to be shown that recurrent architectures can overcome the theoretical limitations of MLPs.
In this short paper, we present the somewhat counter-intuitive result that conventional parametrizations of LSTMs and GRUs are also ``doomed to fail'' at forecasting sequences with trend because they have an \textit{activation bottleneck}.
Roughly speaking, an activation bottleneck is a choke-point in the neural network (NN) through which only a limited amount of information can flow.
Activation bottlenecks are problematic in practice since they prevent NNs from learning non-stationary sequences, e.g., straight lines, random walks, or sequences with trend.

Our contributions are: 1) Introduction of the \textit{activation bottleneck} concept. 2) Analytical exposition of the limitations of NNs with activation bottleneck. 3) Experimental validation of our theoretical findings. 4) Discussion of strategies to mitigate the effect of activation bottlenecks.

Expert readers will observe that many of the following results are straightforward or have already been informally discussed in forums or on the internet. We formalize this ``colloquial knowledge'' and raise awareness for the subject---many practitioners (and scholars) are not aware of the associated limitations of NNs and do not know how to select appropriate hyper-parameters for sequence learning tasks.


\section{Theory}

In this section, we first formalize the notion of activation bottlenecks, and then conduct an analytical exposition of the limits of NNs with activation bottlenecks.
\subsection{Definitions}\label{sec:def}

Let $\R^d$ denote the $d$-dimensional Euclidean space, let $\D^d\subseteq \R^d$ denote a subset of $\R^d$, and let $f:\D^n\rightarrow\D^m$ denote a function to be learned. 
We define a hidden layer to be a function $h: \D^p \rightarrow \D^q$ for arbitrary natural numbers $p,q$.
A hidden layer $h$ has parameters $\bm{\theta}_h=\{W,b,\sigma,\aleph\}$, where $W\in\R^{q\times p}$ are weights, $b\in\R^q$ is a bias term, $\sigma:\R\rightarrow\R$ is a activation function, and $\aleph$ summarizes any further parameters.
Activation function $\sigma$ is sigmoidal if $\underset{t\to-\infty}{\lim} \sigma(t)=c_0$, $\underset{t\to+\infty}{\lim} \sigma(t)=c_1$ for some constants $c_0,c_1\in \R$ and $\sigma(\cdot)$ is non-decreasing everywhere.
A neural network is a function $g:\D^n\rightarrow\D^m$ consisting of a linear input layer $\iota:\D^n\rightarrow\D^{p_1}$, a linear output layer $\omega:\D^{q_k}\rightarrow\D^m$, and $k<\infty $ hidden layers $h_1,\mathellipsis,h_k$ in between s.t. for all $i\in \{1,\mathellipsis,k\}$, the domain of $h_i$ is equal to the co-domain of $h_{i-1}$, and the co-domain of $h_i$ is equal to the domain of $h_{i+1}$. We summarize neural network $g$ and its layers with the following notation
\begin{equation}\label{eq:nn}
    g = \omega\circ h_k \circ \mathellipsis \circ h_1 \circ \iota.
\end{equation}
NN $g$ is Lipschitz continuous if all its hidden layers are Lipschitz continuous since the composition of finitely many Lipschitz continuous functions is Lipschitz continuous~\citep[Theorem 12.6]{eriksson2013applied}.
Note that the characterization in Equation~\eqref{eq:nn} of NNs does not cover networks with skip-connections such as residual NNs~\citep{he2016deep} (sometimes also called \textit{ResNet}).
\begin{definition}[Maximum Approximation Error]\label{def:mae}
    Let $f$ and $g$ be defined as above. The maximum approximation error for $g$ on $f$ is given by
    \begin{equation}
        \varepsilon^\star_{f,g} = \sup \{\varepsilon\in\R^+:\exists x \in \D^n:\lvert\lvert f(x)-g(x)  \rvert\rvert\ \ge \varepsilon\}.
    \end{equation}
\end{definition}
Roughly speaking, the maximum approximation error quantifies the ``biggest'' mistake that a network can make when trained for a specific function.
We will use this notion to demonstrate that certain NN architectures cannot forecast unbounded sequences.

\begin{definition}[Activation Bottleneck]\label{def:ab}
    Neural network $g$ has an activation bottleneck if $g$ has at least one hidden layer $h_i$ whose image is bounded.
\end{definition}
We say that the activation bottleneck is ``located'' in the hidden layer whose image is bounded.

\subsection{Results}

It follows immediately from Definition~\ref{def:ab} that every Lipschitz continuous NN with sigmoidal $\sigma$ in at least one layer has an activation bottleneck since all sigmoidal activation functions have a bounded image.
Note that the class of NNs with activation bottleneck covers many widely-used sequence modeling architectures: LSTMs and GRUs with their default activation functions, as well as MLPs and CNNs with $\tanh$ or logistic/sigmoid activation functions.

Next, let us characterize how an activation bottleneck affects NNs.
\begin{lemma}\label{lemma:ab}
    If neural network $g$ has an activation bottleneck in hidden layer $h_i$ and the layers after $h_i$ are Lipschitz continuous, then the image of $g$ is bounded.
\end{lemma}
\begin{theorem}\label{thm:infinity_error}
    Let $f:\D^n\rightarrow\D^m$ be surjective, and let $\D^n,\D^m$ be unbounded. Then for every neural network $g$ an activation bottleneck in $h_i$ and Lipschitz continuous layers after $h_i$, it holds that
    \begin{equation}\label{eq:infinite_error}
        \varepsilon^\star_{f,g}=\infty.
    \end{equation}
\end{theorem}
In other words, Lemma~\ref{lemma:ab} and Theorem~\ref{thm:infinity_error} state that NNs with an activation bottleneck and ``conventional''\footnote{Almost all widely used NN layers are Lipschitz continuous, e.g., \textit{convolution, normalization, fully-connected, linear, encoder, decoder, softmax}. Note that \textit{dropout} is locally Lipschitz continuous, but not globally.} NN layers afterwards are bounded and hence cannot learn unbounded functions.
The error between prediction and ground truth eventually becomes arbitrarily large.
The class of \textit{learnable functions}~\citep{vapnik1998statistical} for which this behavior occurs includes sequences with trend, random walks, and even straight lines such as the sequence $x_t=x_{t-1}+1$.





\section{Experiments}
\begin{figure}[!t]
    \centering
    \includegraphics[width=0.495\textwidth]{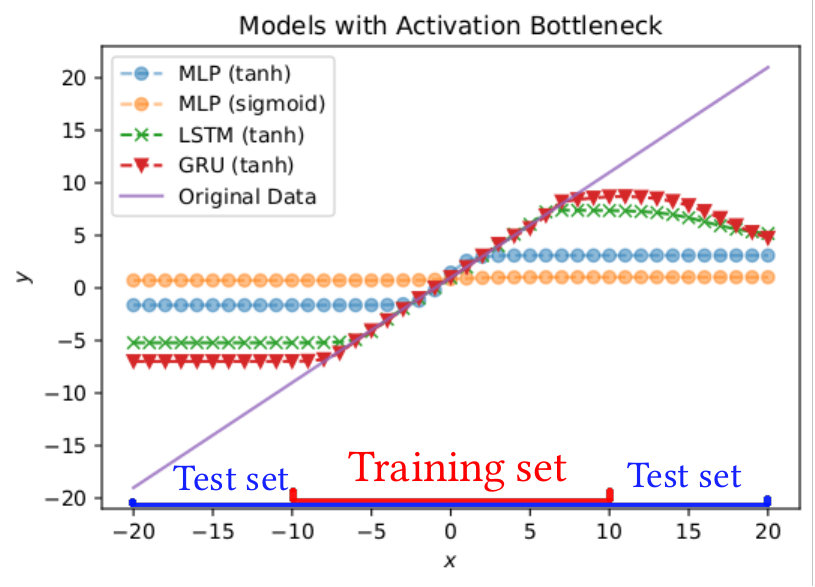}
    \includegraphics[width=0.495\textwidth]{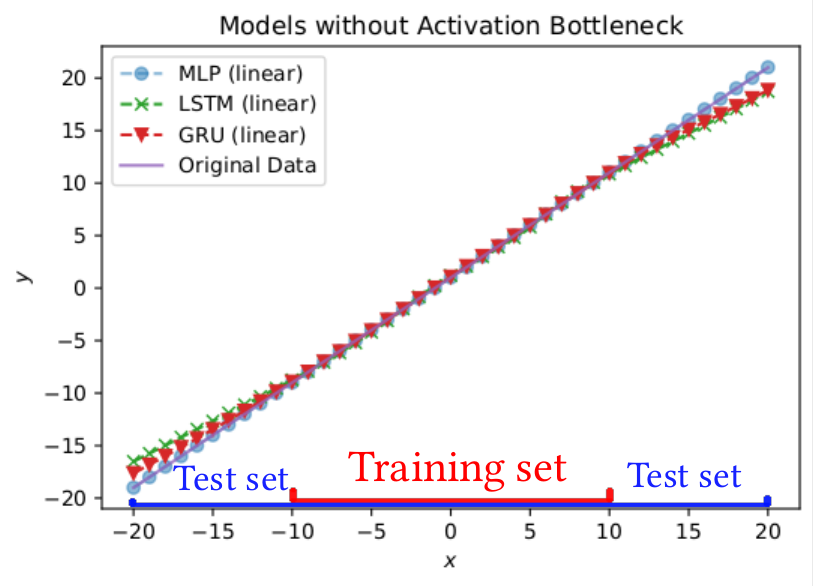}
    \caption[Results of the experiment.]{Results of the experiment. \textit{Left}) Models with activation bottleneck struggle to fit/forecast a simple straight line. \textit{Right}) Models without activation bottleneck do not have this limitation and easily solve this learning task. }
    \label{fig:ab}
\end{figure}

We conduct an experiment to demonstrate the effect of activation bottlenecks.
We compare NN models with activation bottleneck and models without activation bottleneck.
\begin{itemize}
\item Models with activation bottleneck: 1) MLP: two layers of one unit each, $\tanh$ and sigmoid activation, respectively. 2) LSTM: 10 units, standard activation as in Pytorch and Tensorflow. 3) GRU: 10 units, standard activation as in Pytorch and Tensorflow.
\item Models without activation bottleneck: 1) MLP: two layers of one unit each, linear activation. 2) LSTM: 10 units, linear activation (instead of $\tanh$) on recurrent and cell output edges. 3) GRU: 10 units, linear activation (instead of $\tanh$) on recurrent and cell output edges.
\end{itemize}
All models are trained using ADAM and mean squared error loss and 100 training epochs.
Recurrent models (LSTM and GRU) are trained to forecast the next data point based on the previous 10 data points. The MLP is trained similarly, but not in sequential order.
As dataset, we simulate a straight line from $-20$ to $20$ consisting of $41$ data points---this small size is sufficient.
Data within the interval $[-10;10]$ are used as training set, and all data in $[-20;20]$ are used as test set.
We do not use an evaluation measure to quantify the prediction error since we are confident that effect of activation bottlenecks will be visually apparent to most readers.
The results are depicted in Figure~\ref{fig:ab}.

\section{Discussion}
\paragraph{Link to theoretical results.} It is apparent from Figure~\ref{fig:ab} that NNs with activation bottleneck struggle to fit/forecast straight lines.
The same problem arises for any unbounded sequence, since a finite training set is never sufficient to capture the full range of an unbounded sequence.
This implicit ``out-of-distribution'' setting is inherent to all unbounded sequence learning task.
The fact that NNs are universal approximators is of little use here since in the case of finite network width and depth the input domain must be bounded~\citep{guliyev2018approximation}.

\paragraph{How to fix activation bottlenecks.}
The recommended strategy to overcome the effects of an activation bottleneck is to circumvent it, e.g., by introducing skip-connections to the NN which bypass the affected layer.
Thus, the network is given the ability to learn an unbounded sequence component such as a trend without the need to interfere with well-established network parametrizations.
A second strategy is to directly modify the layer where the bottleneck is located to ensure that this layer can output unbounded values.
For most network layers, it is sufficient to exchange sigmoidal activation functions such as $\tanh$ and logistic with ReLU or linear activation.
Note, however, that this strategy may lead to exploding gradients during training.
A third strategy is to introduce a non-Lipschitz continuous layer after the bottleneck, e.g., an inverse sigmoid $\tilde{\sigma}(x)=\log \frac{-x}{x-1}$.
However, this strategy may significantly aggravate finding a good training setup, and will likely lead to numeric instability.

\paragraph{Practical advise.} Unbounded sequences do not exist in finite reality, yet many sequences will eventually leave the limits of a NN's training data, which leads to a similar effect as observed in Figure~\ref{fig:ab}. It is not recommended to try overcoming activation bottlenecks with pre-processing.
MinMax normalization of the training and test set assumes that these data are bounded, which in the case of unbounded sequences has the nasty side effect of seemingly improving the results while the network will quickly struggle with values of the sequence outside of the experimental setup.
Likewise, pre-processing with a function that can project unbounded sequences into a finite range, e.g., a Gaussian kernel, has the downside that one would need infinite floating point precision to avoid numeric instability.
If, despite our warnings, NNs with activation bottleneck are used for potentially unbounded sequences, frequent re-training is recommended to adapt to new data ranges.

To conclude, we remark that the conventional approach of using sigmoidal NN layers backfires for unbounded sequences.
The dogmatic assertion that available data have a similar distribution and/or (auto-)covariance structure as future data is not warranted in all settings where NNs are used.
Time series data such as financial indices or astronomic observations are known to have gradually increasing data ranges, and require different machine learning approaches than language or image data.






\bibliography{activation_bottleneck}

\begin{thebibliography}{8}
\providecommand{\natexlab}[1]{#1}
\providecommand{\url}[1]{\texttt{#1}}
\expandafter\ifx\csname urlstyle\endcsname\relax
  \providecommand{\doi}[1]{doi: #1}\else
  \providecommand{\doi}{doi: \begingroup \urlstyle{rm}\Url}\fi

\bibitem[Cho et~al.(2014)Cho, van Merrienboer, Gulcehre, Bahdanau, Bougares,
  Schwenk, and Bengio]{cho2014learning}
Kyunghyun Cho, Bart van Merrienboer, Caglar Gulcehre, Dzmitry Bahdanau, Fethi
  Bougares, Holger Schwenk, and Yoshua Bengio.
\newblock Learning phrase representations using rnn encoder--decoder for
  statistical machine translation.
\newblock In \emph{Proceedings of the 2014 Conference on Empirical Methods in
  Natural Language Processing (EMNLP)}, pages 1724--1734, 2014.

\bibitem[Eriksson et~al.(2013)Eriksson, Estep, and
  Johnson]{eriksson2013applied}
Kenneth Eriksson, Donald Estep, and Claes Johnson.
\newblock \emph{Applied mathematics: Body and soul: Volume 1: Derivatives and
  geometry in IR3}, volume~1.
\newblock Springer Science \& Business Media, 2013.

\bibitem[Guliyev and Ismailov(2018)]{guliyev2018approximation}
Namig~J Guliyev and Vugar~E Ismailov.
\newblock On the approximation by single hidden layer feedforward neural
  networks with fixed weights.
\newblock \emph{Neural Networks}, 98:\penalty0 296--304, 2018.

\bibitem[He et~al.(2016)He, Zhang, Ren, and Sun]{he2016deep}
Kaiming He, Xiangyu Zhang, Shaoqing Ren, and Jian Sun.
\newblock Deep residual learning for image recognition.
\newblock In \emph{Proceedings of the IEEE conference on computer vision and
  pattern recognition}, pages 770--778, 2016.

\bibitem[Hochreiter(1991)]{hochreiter1991untersuchungen}
Sepp Hochreiter.
\newblock Untersuchungen zu dynamischen neuronalen netzen.
\newblock \emph{Diploma, Technische Universit{\"a}t M{\"u}nchen}, 91\penalty0
  (1):\penalty0 31, 1991.

\bibitem[Hochreiter and Schmidhuber(1997)]{hochreiter1997long}
Sepp Hochreiter and J{\"u}rgen Schmidhuber.
\newblock Long short-term memory.
\newblock \emph{Neural computation}, 9\penalty0 (8):\penalty0 1735--1780, 1997.

\bibitem[Vapnik et~al.(1998)Vapnik, Vapnik, et~al.]{vapnik1998statistical}
Vladimir~Naumovich Vapnik, Vlamimir Vapnik, et~al.
\newblock Statistical learning theory.
\newblock 1998.

\bibitem[Zhang and Qi(2005)]{zhang2005neural}
G~Peter Zhang and Min Qi.
\newblock Neural network forecasting for seasonal and trend time series.
\newblock \emph{European journal of operational research}, 160\penalty0
  (2):\penalty0 501--514, 2005.

\end{thebibliography}
\newpage
\clearpage
\appendix

\section{Proofs}

\subsection{Proof of Lemma~\ref{lemma:ab}}
Recall that Lemma~\ref{lemma:ab} states the following: \textit{If neural network $g$ has an activation bottleneck in hidden layer $h_i$ and the layers after $h_i$ are Lipschitz continuous, then the image of $g$ is bounded.}.
\begin{proof}
    Assume that $g$ has an activation bottleneck. Then $g$ must have at least one hidden layer whose image is bounded. Let $h_i$ be the last such hidden layer in $g$. Observe that all layers of $g$ after $h_i$ are Lipschitz continuous. Hence, in this case the image of $g$ must be bounded since the composition of a function with bounded image and a Lipschitz continuous function is bounded.
\end{proof}

\subsection{Proof of Theorem~\ref{thm:infinity_error}}
Recall that Theorem~\ref{thm:infinity_error} states the following: \textit{Let $f:\D^n\rightarrow\D^m$ be surjective, and let $\D^n,\D^m$ be unbounded. Then for every neural network $g$ an activation bottleneck in $h_i$ and Lipschitz continuous layers after $h_i$, it holds that $\varepsilon^\star_{f,g}=\infty.$}
\begin{proof}
Let $f$ be defined as in Theorem~\ref{thm:infinity_error}.
Since $g$ has an activation bottleneck, the image of $g$ must be bounded due to  Lemma~\ref{lemma:ab}.
Hence, there exists a finite bound $\varepsilon > 0$ such that for every $u_1,v_1\in\D^n$ we have $\lvert\lvert g(u) - g(v) \rvert\rvert < \varepsilon$.
However, since $\D^m$ is unbounded and $f$ is surjective, for every finite $\varepsilon$ there exist $u_2,v_2\in \D^n$ such that $\lvert\lvert f(u) - f(v) \rvert\rvert \ge \varepsilon$.
Since $u_1,u_2,v_1,v_2$ are all elements of the same set $\D^n$, it follows that the set $$\mathcal{E}=\{\varepsilon \in \R^+:\exists x \in \D^n: \lvert\lvert f(x) - g(x) \rvert\rvert \ge \varepsilon\}$$
must be equal to $\R^+$.
Since the maximum approximation error $\varepsilon^\star_{f,g}$ as per Definition~\ref{def:mae} is the supremum of $\mathcal{E}=\R^+$, we can conclude that Equation~\ref{eq:infinite_error} is correct, i.e., $\varepsilon^\star_{f,g}=\infty$.

\end{proof}

\end{document}